\documentstyle[11pt,fullpage,psfig]{article}

\newcommand {\GIG} {Graph Interpolation Grammar}

\newcommand {\GIGs} {Graph Interpolation Grammars}

\newcommand {\gils} {graph interpolation languages}

\newcommand {\GILs} {Graph Interpolation Languages}

\newcommand {\paper} {report}

\newcommand {\tuple}[1] {\langle #1 \rangle}
\newcommand {\length}[1] {|#1|}

\newcommand {\union} {\cup}
\newcommand {\inter} {\cap}
\newcommand {\gigpath}[2] {\mbox{~$\begin{array}{c}
    \stackrel{#1}{\Longrightarrow} \\[-5pt]
    \scriptscriptstyle \rm #2
  \end{array}$~}}
\newcommand {\pipe} {\mid}
\newcommand {\derives}[1] {\mbox{~$\begin{array}[t]{c}
    \Longrightarrow \\[-8pt]
    \scriptscriptstyle \rm #1
  \end{array}$~}}
\newcommand {\mderives}[1] {\mbox{~$\begin{array}[c]{c}
    * \\[-8pt]
    \Longrightarrow \\[-8pt]
    \scriptscriptstyle \rm #1
  \end{array}$~}}
\newcommand {\pderives}[1] {\mbox{~$\begin{array}[c]{c}
    + \\[-8pt]
    \Longrightarrow \\[-8pt]
    \scriptscriptstyle \rm #1
  \end{array}$~}}

\newcommand {\subsubsubsection}[1] {\paragraph{#1}}

\long\def\startIgnoring #1\stopIgnoring{}
\def\stopIgnoring{}

\newtheorem{theorem}{Theorem}
\newtheorem{lemma}{Lemma}

\newenvironment{sketch}{{\bf Proof outline}}{\mbox{ $\Box$}\\ 
}
\newenvironment{example}{{\bf Example}}{\mbox{ $\diamond$}\\ 
}

\startIgnoring
\RRtitle{Etude des Grammaires à Interpolations de Graphes \\
en tant qu'Automates Non Contextuels}

\RRetitle{Graph Interpolation Grammars as Context-Free Automata}

\titlehead{Graph Interpolation Grammars as CF Automata}

\RRauthor{John Larchevêque}

\RRdate{July 1998}

\RRtheme{3}
\RRprojet{Atoll}

\URRocq

\RRabstract{A derivation step in a Graph Interpolation Grammar
has the effect of scanning an input token. This feature, which
aims at emulating the incrementality of the natural parser,
restricts the formal power of GIGs. This contrasts with the fact
that the derivation mechanism involves a context-sensitive device
similar to tree adjunction in TAGs. The combined effect of
input-driven derivation and restricted context-sensitiveness
would be conceivably unfortunate if it turned out that Graph
Interpolation Languages did not subsume Context Free Languages
while being partially context-sensitive. This report sets about
examining relations between CFGs and GIGs, and shows that GILs
are a proper superclass of CFLs. It also brings out a strong
equivalence between CFGs and GIGs for the class of CFLs. Thus, it
lays the basis for meaningfully investigating the amount of
context-sensitiveness supported by GIGs, but leaves this
investigation for further research.}

\RRkeyword{parsing, natural language processing, linguistics, grammar, syntactic representation}

\RRresume{Une étape de dérivation dans une Grammaire à
Interpolations de Graphes a pour effet d'intégrer un lexème de la
chaîne d'entrée. Cette propriété, qui vise à émuler
l'incrémentalité de l'analyseur syntaxique naturel, restreint la
puissance formelle des GIG.  Ceci est à contraster avec le fait
que la dérivation d'une chaîne par une GIG met en jeu un
mécanisme à effet contextuel analogue à l'adjonction d'arbre dans
une TAG. L'effet conjoint du comportement d'automate des
dérivations et d'une certaine mesure de contextualité pourrait
aboutir à une situation potentiellement fâcheuse dans laquelle
les languages à interpolations de graphes ne subsumeraient pas
les languages non contextuels, tout en étant partiellement
contextuels. Ce rapport examine les relations entre les CFG et
les GIG, et montre que les langages à interpolations de graphes
constituent une surclasse stricte des langages non
contextuels. Il y est également montré que les GIG sont fortement
équivalentes aux CFG pour la classe des languages non
contextuels. Ces résultats donnent un sens à la question du degré
de contextualité des GIG, qui pourra ainsi faire l'objet de
travaux futurs.}

\RRmotcle{syntaxe, langage naturel, linguistique, grammaire, représentation syntaxique}
\stopIgnoring

\begin{document}

\title{\GIGs\ as context-free automata}
\author{John Larchev\^{e}que}
\date{July 1998}
\maketitle
\abstract{A derivation step in a Graph Interpolation Grammar
has the effect of scanning an input token. This feature, which
aims at emulating the incrementality of the natural parser,
restricts the formal power of GIGs. This contrasts with the fact
that the derivation mechanism involves a context-sensitive device
similar to tree adjunction in TAGs. The combined effect of
input-driven derivation and restricted context-sensitiveness
would be conceivably unfortunate if it turned out that Graph
Interpolation Languages did not subsume Context Free Languages
while being partially context-sensitive. This report sets about
examining relations between CFGs and GIGs, and shows that GILs
are a proper superclass of CFLs. It also brings out a strong
equivalence between CFGs and GIGs for the class of CFLs. Thus, it
lays the basis for meaningfully investigating the amount of
context-sensitiveness supported by GIGs, but leaves this
investigation for further research.}


\section{Introduction}

\subsection{\GIGs\ as automata}

\GIGs\ are indifferently viewed as grammars or automata. That is to
say that a derivation step in a GIG is an automaton move during
which one token of the input string is scanned. In this
perspective, GIG rules can be viewed as the finite control of an
automaton.

\subsection{Rationale for GIG graphs}

The output of a GIG is a graph rather than a tree. If trimmed
down to its bare essentials, a GIG graph can be considered as a
particular representation of a tree, in which a node is connected
to its leftmost child through a {\em parent-of} edge, and each
child is connected to its right sibling through a {\em
predecessor-of} edge. Now, the advantages of this representation
are the following
\begin{enumerate}
\item A parse subgraph is allowed to assign a smaller arity to a
given node than the complete graph that includes it, so that
arities can be determined incrementally. For example, given a
sentence starting with \mbox{\em John thinks Mary \ldots}, what
can be hypothesized is that {\em Mary} is the subject of a verb,
but the arity of this verb is unknown at this stage of the
parsing.
\item In addition to tree adjunction, which shifts a subtree further
down the parse tree, parse graphs support a similar operation
which shifts a subgraph further right, thus allowing sets of
constructions such as the following to be viewed as
paradigms. (Alternating paradigm constituents appear in
square brackets and gaps are represented by squares).

\begin{quote}
Who did you [invite] $\Box$\/? \\
Who did you [happen to invite] $\Box$\/? \\
Who did you [think I invited] $\Box$\/?
\end{quote} 

\end{enumerate}

Such a paradigm is captured in GIG by considering that the basic
construction (with the constituent {\em invite}) can be expanded
by interpolating a subgraph to the left of the gap. One can
expect it to be hard, if possible at all, to represent such a
constituent as \mbox{\em think I invited} (without the object
gap) by a tree, especially if intuitive phrase structure is to be
preserved. Hence the need for a more versatile structure than
standard parse trees.

\subsection{Expressive power of graph interpolation}

The operation on which GIGs are based, graph interpolation, is
basically a tree adjunction~\cite{joshi-etal-75,Vijay94} applied
to a pair of GIG graphs. On the one hand, graph interpolation is
more general than tree adjunction, but, on the other hand, its
application obeys a very strong constraint, namely that it be
performed in response to the scanning of a token. Owing to this
constraint, it is even legitimate to ask whether Graph
Interpolation Languages subsume Context-Free Languages. So,
before solving the important problem of GIGs relation to TAGs in
term of formal power, it is worth comparing GIGs with CFGs. This
comparison is the subject of the present \paper.

\subsection{Content of the \paper}

This \paper\ proves that 
\begin{itemize}
\item Graph Interpolation Languages include Context-Free
Languages (Section~\ref{inclusion.sec}),
\item Graph Interpolation Languages properly include Context-Free
Languages (Section~\ref{properInclusion.sec}),
\item GIGs are strongly equivalent to CFGs on the class of CFLs
(Section~\ref{strongEquivalence.sec}),
\item the intersection of CFLs and deterministic GILs includes
the class LL(1) (Section~\ref{determinism.sec}).
\end{itemize}

A preliminary to these results is the definition of a GIG, which
is given in the next section.

\section{Definition of a \GIG}

A linguistic presentation of \GIGs\ was given in \cite{larcheveque98a}. 
The definition given here deliberately ignores all features not
directly relevant to the purpose of defining GIGs as automata.
\begin{itemize}
\item Multiple-interpolation rules will not be considered, for the
formal power they contribute is excessive in the perspective of a
comparison with CFGs.
\item A fixed word order will be assumed.
\item Phrase heads and grammatical functions will be ignored, which
shows that they can be decoupled from parsing issues.
\item Node label subtyping, which has no impact on formal power,
will not be made use of.
\end{itemize}

Apart from these introductory remarks, the present \paper\ can be
read without any prior knowledge of \GIGs.

\subsection{Parse graphs}

A parse graph is a representation of a tree in which the relation
{\em parent-of} is broken down into a {\em first-child} edge and
zero or more {\em right-sibling} edges. More precisely, a parse
graph $\Gamma$ is a
tuple $\tuple{N, L, r, V, \Sigma, \gamma, E, F}$, where
\begin{itemize}
\item $N$ and $L$ are disjoint sets of nodes (respectively
nonlexical nodes and lexemes),
\item $r$ is a distinguished member of $N$, called the graph {\em root},
\item $V$ and $\Sigma$ are disjoint sets of node labels
(respectively variables and symbols),
\item $\gamma$ is the label assignment, i.e. a function in $N
\times V \union L \times \Sigma$,
\item $E$ is the set of {\em first-child} edges, i.e. a function
in $N \times (N \union L)$,
\item $F$ is the set of {\em right-sibling} edges, i.e. a
function in $N \times N$. (This definition precludes lexical
nodes, i.e. members of $L$, from having siblings.)
\end{itemize}

\subsubsection{Constraints on $E$ and $F$}
\label{graphShape.sec}

\sloppy
\begin{itemize}
\item $E$ and $F$ are functions, i.e. $\length{E(n)} \leq 1$ and
$\length{F(n)} \leq 1$ for all $n$ in $N$. (The notation ${\cal
R}(x)$, where $\cal R$ is a relation, denotes the set $\{
y~|~\tuple{x,y} \in {\cal R}\}$.) In a {\em complete parse
graph}, $N$ is the domain of $E$, i.e. $\length{E(n)} = 1$ for
all $n$ in $N$.
\item Every node except the root has exactly one immediate
dominator\footnote
{A node $n$ is said to {\em dominate} a node $m$ in a rooted
graph whenever every path from the root to $m$ goes through $n$. The
term {\em dominator} is here preferred to {\em predecessor} or
{\em successor} to avoid confusions with the {\em successor}
relation introduced in Section~\ref{successor.sec}.}
by $E \union F$, i.e. \mbox{$\length{(E
\union F)^{-1}(n)} = 1$} for all $n$ in $N - \{ r \} \union L$.
\item Every node in $L$ is a single child with no descendant,
i.e. $E(l) = F(l) = F^{-1}(l) = \emptyset$ for all $l$ in $L$. 
\end{itemize}

\subsection{Path interpolation}

A path interpolation consists in plugging into a parse graph $\Gamma_1
= \tuple{N_1, L_1, r_1, V, \Sigma, \gamma_1, E_1, F_1}$ a
disjoint parse graph $\Gamma_2 =
\tuple{N_2, L_2, s, V, \Sigma, \gamma_2, E_2, F_2}$ by
substituting a path of $\Gamma_2$ for a node $q$ of $\Gamma_1$ that is
called the {\em anchor} of the interpolation. The path that is
substituted is defined by 3~nodes of $\Gamma_2$, $s$, the source of
the path (which is the root of $\Gamma_2$), $t$, the target of the path,
and $s'$, the successor of $s$ in the interpolation
path\footnote
{Examples of interpolations are shown on Figure~\ref{shifts.fig}.}.

Owing to properties of $E_2$ and $F_2$ (i.e. existence of at most
one immediate dominator for any node), there is at most one path
from $s$ to $t$. A constraint on path interpolation is that there
be exactly one path (of length~0 or more) from $s$ to $t$.

There are 3~types of interpolation.
\begin{itemize}
\item In a {\em down-shift}, $\tuple{s,s'}$ is in $E_2$. A
down-shift has the effect of a tree adjunction.
\item In a {\em right-shift}, $\tuple{s, s'}$ is in $F_2$. A
right-shift has the effect of shifting a sibling edge of $\Gamma_1$
further right in the resulting graph.
\item In an {\em $\epsilon$-shift}, $s=s'=t$. The effect of an
$\epsilon$-shift is to insert a subgraph at the anchor~$q$.
\end{itemize}

Formally, an interpolation $\cal I$ is a function that maps a
tuple of the form $\tuple{\Gamma_1, \Gamma_2, q, t}$ to a parse
graph $\Gamma_3$. ($q$ is the anchor of the interpolation, and
$t$ the target, i.e. the endpoint, of the interpolation path.)
Let $\Gamma_1$ and $\Gamma_2$ be defined as above, then $\Gamma_3 = 
\tuple{N_3, L_3, r_3, V, \Sigma, \gamma_3, E_3, F_3}$ is defined
as follows, where $s$ is the root of $\Gamma_2$ (i.e. the interpolation
source) and $s'$ is the node that immediately follows $s$ in the
interpolation path.
\begin{itemize}
\item $N_3 = N_1 - \{ q \} \union N_2$
\item $L_3 = L_1 \union L_2$
\item $r_3 = r_1$
\item $\gamma_3 = \gamma_1 \union \gamma_2$

\item \begin{displaymath} \begin{array}{llll}
E_3 = & & E_1 \union E_2 \\
\multicolumn{4}{l}{\mbox{\it // parent of q, if any, becomes parent of s}} \\
& -     & \{ \tuple{n,q}~| &n \in N_1 \} \\
& \union & \{ \tuple{n, s}~| &\tuple{n, q} \in E_1 \} \\
\multicolumn{4}{l}{\mbox{\it // child of q, if any, becomes child of t (down-shift) or s (right-shift)}} \\
& -     & \{ \tuple{q,n}~| &n \in N_1 \} \\
& \union & \{ \tuple{m,n}~| & \tuple{q,n} \in E_1 \mbox{~and~} \\
&        &                   & ((\tuple{s,s'} \in E_2 \mbox{~and~} m=t)  \\
&        &                   & \mbox{~or~} (\tuple{s,s'} \notin E_2 \mbox{~and~} m=s)) \} 
\end{array} \end{displaymath} 

\item \begin{displaymath} \begin{array}{llll}
F_3 = & & F_1 \union F_2 \\
\multicolumn{4}{l}{\mbox{\it // left sibling of q, if any, becomes left sibling of s}} \\
& -     & \{ \tuple{n,q}~| &n \in N_1 \} \\
& \union & \{ \tuple{n, s}~| &\tuple{n, q} \in F_1 \} \\
\multicolumn{4}{l}{\mbox{\it // right sibling of q, if any, becomes right sibling of t (right-shift) or s (down-shift)}} \\
& -     & \{ \tuple{q,n}~| &n \in N_1 \} \\
& \union & \{ \tuple{m,n}~| & \tuple{q,n} \in F_1 \mbox{~and~} \\
&        &                   & ((\tuple{s,s'} \in F_2 \mbox{~and~} m=t)  \\
&        &                   & \mbox{~or~} (\tuple{s,s'} \notin F_2 \mbox{~and~} m=s)) \} 
\end{array} \end{displaymath} 
\end{itemize}

\subsubsection{Definition constraints for $\cal I$}

\begin{enumerate} 
\item \label{disjointGraphs.con}
{\bf Disjunction constraint}
$\Gamma_1$ and $\Gamma_2$ are disjoint, i.e. $N_1 \inter N_2 =
\emptyset$ (from which it follows that the other pairs of sets
are disjoint too, given constraints on parse graph edges).
\item \label{monoLexicalized.con}
{\bf Lexicalization constraint}
$\Gamma_2$ is mono-lexicalized, i.e. $\length{L_2}=1$.
\item \label{leftmostLexicalized.con}
{\bf Context constraint}
$\Gamma_1$ is leftmost-lexicalized, meaning that its frontier,
i.e. its yield over $(N_1 \union L_1)^*$, is in the regular language
$L_1^*N_1^*$. (The concept of frontier is defined more precisely in
the next section.)
\item \label{frontier.con}
{\bf Frontier constraint}
Given $x\alpha$ the frontier of $\Gamma_1$, with $x$ in $L_1^*$ and
$\alpha$ in $N_1^*$, and $\delta a\beta$ the frontier of $\Gamma_2$, with $a$
in $L_2$ and $\delta$ and $\beta$ in $N_2^*$, the frontier of $\Gamma_3$ has prefix
$xa$\footnote
{For details on the possible forms of the frontier of $\Gamma_3$, see
Section~\ref{determinism.sec}.}.
\end{enumerate}

\begin{figure}[htbp]
\begin{center} ~\psfig{file=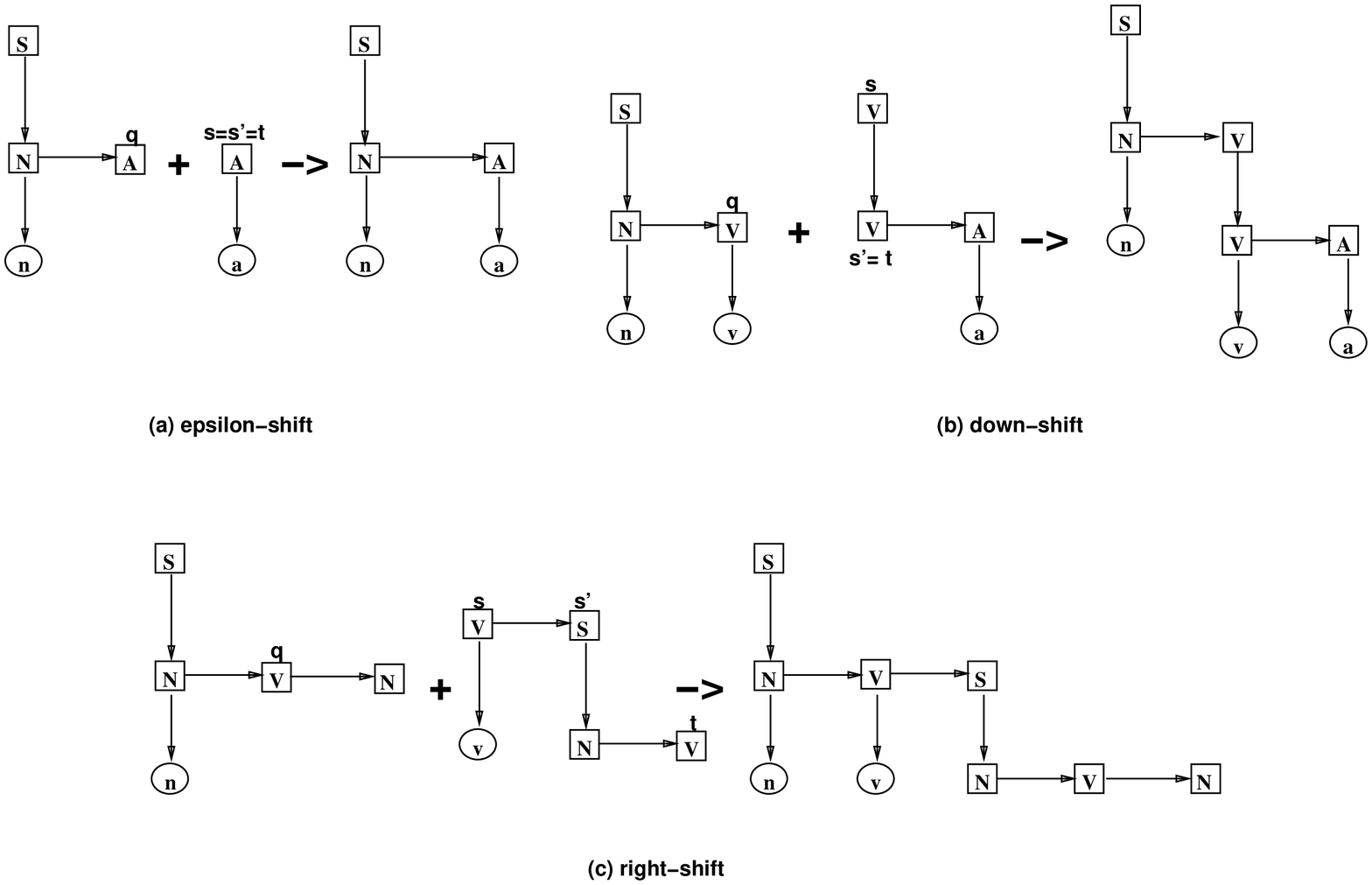,width=\textwidth} \end{center}
\caption{The three types of interpolation}
\label{shifts.fig}
\end{figure}

Figure~\ref{shifts.fig} illustrates the three forms of
interpolation.

\subsubsection{Anchor position}
\label{successor.sec}

Given a triple $\tuple{\Gamma_1,\Gamma_2,t}$, the constraints just
enumerated restrict the set of possible positions for the anchor
$q$ with respect to the last lexeme of $\Gamma_1$ if
$\tuple{\Gamma_1,\Gamma_2,q,t}$ is to belong to the definition domain of
interpolation. As a matter of fact, $q$ is either an ancestor or
an immediate successor of the last lexeme of $\Gamma_1$, as this
section will attempt to prove\footnote
{This result does not imply that there is a single choice of $q$
for given $\Gamma_1$, $\Gamma_2$, and $t$.}.

\subsubsubsection{Immediate successor}

The notion of ``immediate successor'' is defined as follows. Let
$<$ be the ``successor'' relation in a parse graph $\Gamma$, defined
as follows.  (The operator $\circ$ denotes relation composition
or, equivalently, edge concatenation.) 

\begin{displaymath} \begin{array}{lll}
< & = & ((F^{-1})^* \circ E^{-1})^* \circ F \circ E^*
\end{array} \end{displaymath} 

Then, the ``immediate successor'' relation, noted as $\ll$, holds
between $m$ and $n$ if and only if $m < p < n$ implies that $m=p$
or $n=p$.

\subsubsubsection{Graph frontier}

\sloppy
This relation makes it possible to define the frontier of a parse
graph \mbox{$\Gamma=\tuple{N, L, r, V, \Sigma, \gamma, E, F}$} more precisely
than was done in the previous section.
Let $T$ equal \mbox{$\{ n \in N \union L ~|~ E(n) = \emptyset \}$} in:

\begin{displaymath} \begin{array}{lll}
\mbox{frontier}(\Gamma) = \{ w ~| & w \in T^* \mbox{~and~} \\
& \forall x,y \in T^*, \forall m,n \in T: w = xmny \Rightarrow m \ll n \}
\end{array} \end{displaymath}

The frontier of a parse graph is a string containing all of and
only its terminal nodes in $\ll$~order. (The definition of $\ll$
implies that \mbox{$\length{w}=\length{T}$}.)

\subsubsubsection{Anchor position with respect to the last lexeme}

Let $\mu$ be the last lexeme of $\Gamma_1$
and $\nu$ the single lexeme of $\Gamma_2$. It will now be shown that
the constraint $\mu \ll \nu$ (which results from the
frontier constraint) can only be satisfied if $q$,
the anchor, is an ancestor or immediate successor of $\mu$. 
This theorem does not cover the initial interpolation step, which
starts from a context $\Gamma_1$ that consists of a single nonterminal
node (Section~\ref{initialStep.sec}).

\begin{theorem}
If $L_1$ is nonempty, $\cal I$ is defined for $\tuple{\Gamma_1, \Gamma_2,
q, t}$ only if $q$ is either an ancestor or an immediate
successor of the last lexeme of $\Gamma_1$. I.e., let $\mu$ be the
rightmost lexeme in the frontier of $\Gamma_1$, either \mbox{$q \in
(F_1^{-1*} \circ E_1^{-1})^*(\mu)$} or \mbox{$\mu \ll q$} must
hold.
\end{theorem}

\begin{sketch}
The constraint $\mu \ll \nu$ can be decomposed as follows,
where $m$ and $n$ are nodes of $\Gamma_3$. 

\begin{displaymath} \begin{array}{lll}
\mu \gigpath{*}{F_3^{-1*} \circ E_3^{-1}} m \gigpath{\mbox{}}{F_3} n \gigpath{*}{E_3} \nu
\end{array} \end{displaymath}

In order to locate $q$ with respect to $\mu$, we will locate $s$,
the root of $\Gamma_2$, and then use the fact that $s$ occupies in
$\Gamma_3$ the position that $q$ occupies in $\Gamma_1$. More precisely,
the unique $E_1 \union F_1$-path from $r_1$ to $q$ contains the
same nodes (down to $q$ itself) as the unique $E_3 \union
F_3$-path from $r_3$ to $s$.

Now, as the root of $\Gamma_2$, $s$ dominates $\nu$, which implies
that it lies somewhere on a path containing $r_3$, $m$, $n$, and
$\nu$ (in this order), since every node has a single immediate
dominator by \mbox{$E_3 \union F_3$}.

Supposing $s$ properly dominates $m$, then $m = t$, for the path
from $s$ to $m$ occupies the place of an $N_1$ node, namely $q$,
on the path from $r_1$ to $\mu$. (By definition of an
interpolation, $r_1=r_3$.) Conversely, $q$ occupies the place of
$m$ in $\Gamma_1$; therefore {\bf $q$ is an ancestor of $\mu$}.

The same conclusion can be made if $s$ equals $m$.

If $s$ is dominated by $n$, we have $\mu \ll s$, from which
it follows that $\mu \ll q$ in $\Gamma_1$. Therefore, {\bf $q$ is an
immediate successor of $\mu$.}
\end{sketch}

\subsection{Interpolation rule}

An interpolation rule $\rho$ is a tuple \mbox{$\tuple{A, \Gamma_2,
t}$}, where $A$ is a variable in $V$ and $t$, the {\em
interpolation target}, is a node of $\Gamma_2$. Rule $\rho$ is said to
match a parse graph $\Gamma_1$ if the interpolation function is
defined for \mbox{$\tuple{\Gamma_1, \Gamma_2, q, t}$} for some $q$ in
$\gamma_1^{-1}(A)$.

The variable~$A$ is called the {\em context pattern} of the rule,
$\Gamma_2$ is called the {\em addendum}, and a matching
$\Gamma_1$ the {\em context} to which the rule is applied.

\subsubsection{Rule notation}

The clearest way of representing an interpolation involves a
2-dimensional representation of the addendum, as on
Figure~\ref{shifts.fig}. On the other hand, since a parse graph
is a representation of a parse tree, which is itself a
representation of a context-free derivation, an addendum can be represented
as a CF~derivation. Thus, a rule \mbox{$\tuple{A, \Gamma_2, t}$} can
be represented as a left-hand side consisting of the context
pattern $A$, a separating right arrow, and a right-hand side in
which $\Gamma_2$ is represented as a leftmost derivation, and $t$ is
dotted. For example, the $\epsilon$-shift on
Figure~\ref{shifts.fig} can be noted as
\mbox{$A
\rightarrow (\dot{A} \derives{lm} a)$}. Likewise, the down-shift can
be noted as \mbox{$V \rightarrow (V \derives{lm} \dot{V}A \derives{lm}
Va)$} and the right-shift as \mbox{$V \rightarrow (VS \derives{lm}
vS \derives{lm} vN\dot{V})$}.

\subsection{\GIG}
\label{initialStep.sec}

A \GIG\ is a tuple $\tuple{V, \Sigma, P, S}$, where $P$ is a set
of interpolations rules, $V$ and $\Sigma$ are the label sets
used by these rules, and $S$ is a member of $V$ called the {\em axiom}.

At the start of a GIG derivation, the {\em context} is a subgraph
consisting of one node labeled with the axiom.

Each step matches a rule of $P$ with the context and applies the
corresponding interpolation, yielding a new context.

\section{\GILs\ form a superclass of CFLs}
\label{inclusion.sec}

\subsection{A definition of \GILs}

Let an Instantaneous Description be defined as a
tuple~$\tuple{\Gamma,w}$, where the string to parse is of the form
$vw$, with $v$ the yield of the context $\Gamma$. Initially, $\Gamma$
consists of a single node labelled with the axiom, and $v$ is
empty. An interpolation \mbox{${\cal I}(\tuple{\Gamma,\Gamma_2,q,t})=\Gamma_3$}
maps $\tuple{\Gamma,ax}$ to $\tuple{\Gamma_3,x}$, with \mbox{$\mbox{\em
yield}(\Gamma_3)=\mbox{\em yield}(\Gamma).a$}. A string $w$ is accepted by a GIG
if applicable interpolations map the initial ID to
$\tuple{\Gamma_f,\epsilon}$, with \mbox{$\mbox{\em yield}(\Gamma_f)=w$}. The
definition of a Graph Interpolation Language follows in the
classical way.

\subsection{\GILs\ as a superclass of Context-Free Languages}

\begin{theorem}
Any CFL not containing the string~$\epsilon$ can be recognized by
a \GIG.
\end{theorem}

\begin{sketch} 
Any CFL not containing the string~$\epsilon$ can be described by
a grammar in Greibach Normal Form, i.e. a CFG $\tuple{V, \Sigma,
P, S}$ in which every
production is of the form $A \rightarrow a\gamma$, with $A$ in
$V$, $a$ in $\Sigma$, and $\gamma$ in $V^*$.

To build a GIG $G'$ from a GNF grammar~$G$, consider each
production $p$~as a GIG rule~$\rho$ whose context pattern
consists of the left-hand side of $p$, and whose addendum
represents the subtree obtained in a CFG-derivation by
substituting the right-hand side for the left-hand side, except
that the leftmost symbol of the right-hand side is replaced by
the preterminal symbol $X$, \mbox{$X \notin V \union \Sigma$}, as
shown in Figure~\ref{GNFRule.fig}.

\begin{figure}[htbp]
\begin{center} ~\psfig{file=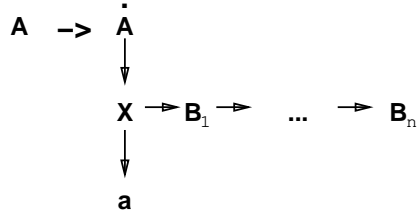,height=33mm} \end{center}
\caption{GIG rule built from the production $A \rightarrow aB_1...B_n$}
\label{GNFRule.fig}
\end{figure}

The GIG thus built is a tuple $\tuple{V', \Sigma, P', S}$, where
$V'$ equals \mbox{$V \union \{ X \}$} and $P'$ is the set of
GIG-rules just described.

It is easy to prove by induction that the derivation produced by
this GIG is a leftmost derivation of $G$.

Indeed, initially, the context consists of a node labeled $S$,
whose yield is therefore a leftmost sentential form of $G$.

Supposing the yield of the current context is a leftmost
sentential form $xA_1...A_n$, $x$ in $\Sigma^*$, then there will
be an applicable GIG rule if and only if there is a production
\mbox{$A_1
\rightarrow aB_1...B_m$} in $P$, and this rule will map the
context to a context whose yield is $xaB_1...B_mA_2...A_n$,
i.e. a leftmost sentential form of~$G$.
\end{sketch}

\section{GILs form a proper superclass of CFLs}
\label{properInclusion.sec}

\begin{theorem}
The class of Context-Free Languages is properly included in the
class of \GILs.
\end{theorem}

\begin{figure}[htbp]
\begin{center} ~\psfig{file=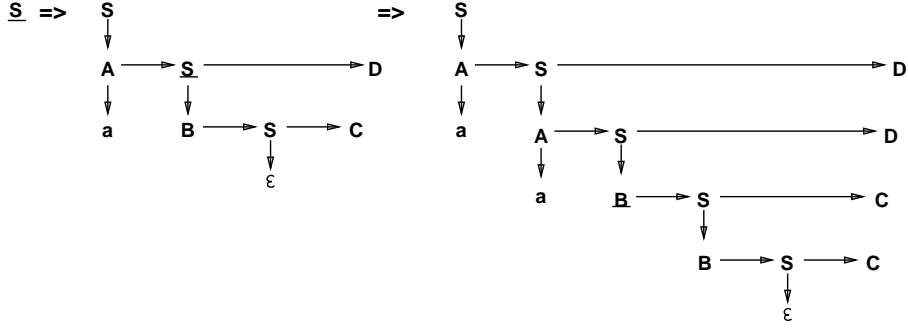,height=45mm} \end{center}
\caption{Initial GIG derivation steps for $aabbccdd$}
\label{csDeriv.fig}
\end{figure}

\begin{sketch} 
Now it has been shown that every CFL not containing~$\epsilon$ is
recognized by some GIG, it remains to show that there are
languages which can be recognized by \GIGs\ but not by
Context-Free Grammars.

The language $a^nb^nc^nd^n$, which is a classic example of a
context-sensitive language, is recognized by the GIG \mbox{$\tuple{\{
S,A,B,C,D \}, \{ a,b,c,d \}, P, S}$}, where $P$ consists of the
following rules:
\begin{displaymath} \begin{array}{llll}
(1)&S&\rightarrow&(\dot{\underline{S}} \Rightarrow \underline{A}SD \Rightarrow a\underline{S}D \Rightarrow aB\underline{S}CD \Rightarrow aB\epsilon CD)\\
(2) & S & \rightarrow &(\underline{S} \Rightarrow \underline{A}SD \Rightarrow a\underline{S}D \Rightarrow aB\dot{S}CD) \\
(3) & B & \rightarrow & (\dot{\underline{B}} \Rightarrow b) \\
(4) & C & \rightarrow & (\dot{\underline{B}} \Rightarrow c) \\
(5) & D & \rightarrow & (\dot{\underline{B}} \Rightarrow d) \\
\end{array} \end{displaymath}

A 2-dimensional representation of these rules appears on
Figure~\ref{csRules.fig}. 

\begin{figure}[htbp]
\begin{center} ~\psfig{file=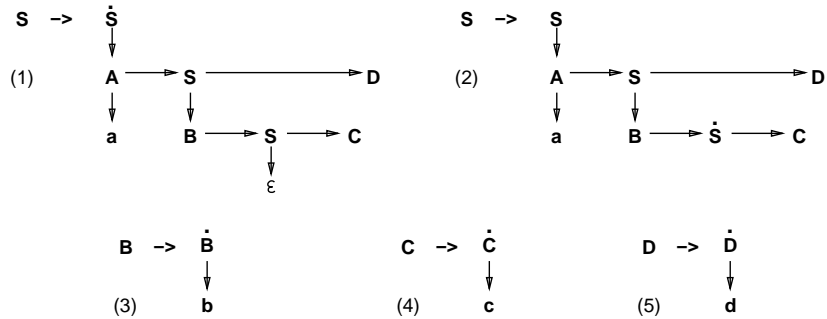,height=45mm} \end{center}
\caption{GIG rules for $a^nb^nc^nd^n$}
\label{csRules.fig}
\end{figure}

As production~(1) requires the presence of an $S$-node in the
context frontier, it applies exclusively to the initial
context. On the other hand, the frontier constraint requires that
the down-shift~(2) apply exclusively to the middle~$S$ of the previous
interpolation, i.e. the $S$ that is an immediate successor (by
$\ll$) of the last lexeme. Figure~\ref{csDeriv.fig} shows the
first two~steps of the unique derivation of~$aabbccdd$.
\end{sketch} 

\section{GILs strongly subsume CFLs}
\label{strongEquivalence.sec}

Define a {\em 2-tiered} Context-Free Grammar to be a CFG
containing no production of the form \mbox{$A \rightarrow \alpha
a\beta$} in which $\alpha$ or $\beta$ is a nonempty string in
\mbox{$(V \union \Sigma)^*$} and $a$ is a
terminal symbol. The graph representations of trees generated by
a 2-tiered CFG obey the constraint that lexical nodes have
neither descendants nor siblings
(Section~\ref{graphShape.sec}). A simple way of making an
arbitrary CFG \mbox{$G=\tuple{V,\Sigma,P,S}$} 2-tiered is to {\em
(i)}~add into $V$ a preterminal symbol $X_a$ for each $a$ in
$\Sigma$, {\em (ii)}~substitute $X_a$ for $a$ in all productions
of $P$, and {\em (iii)}~add into $P$ the production $X_a
\rightarrow a$ for each~$a$ in $\Sigma$. The resulting CFG $G'$
remains close enough to the original grammar $G$ for any semantic
computation on a $G$~tree to map straightforwardly to its
counterpart $G'$~tree, and vice versa. Thus, conversion to
2-tiered form is semantically neutral, and it seems therefore
legitimate to say that GIGs are strongly equivalent to CFGs on
the class of CFLs, even though the derivations obtained with a
GIG are necessarily 2-tiered. In this sense, the following
theorem asserts the strong equivalence of GIGs and CFGs on the
class of CFLs.

\begin{theorem}
Any unambiguous 2-tiered CFG not recognizing the empty string can
be mapped to a GIG that constructs exactly the same
CF~derivations on the same inputs.
\end{theorem}

\begin{sketch}
Consider the following procedure for building a GIG
\mbox{$G'=\tuple{V,\Sigma,P',S}$} from a 2-tiered CFG
\mbox{$G=\tuple{V,\Sigma,P,S}$}: 

\begin{enumerate}
\item For each variable $A$ in $V$, unwind all possible
leftmost derivations by grammar $G$ starting at $A$ until either
{\em (i)}~the empty string is derived (\mbox{$A \pderives{lm}
\epsilon$}), {\em (ii)}~the derived string has a leftmost
terminal symbol (\mbox{$A \pderives{lm} a\alpha$}), or {\em
(iii)}~the leftmost symbol of the derived string is a variable of
$V$ that has already been replaced in the derivation (\mbox{$A
\mderives{lm} B\alpha \pderives{lm} B\beta \alpha$}). Let
$\Delta$ be the set of derivations thus obtained.

Note that, when a leftmost $\epsilon$ is obtained, the derivation
is pursued, if possible, until the leftmost symbol is a member of
$V \union \Sigma$.

These bounded derivations will be used later on to form addenda in GIG rules.

\item For each derivation $D_1$ in $\Delta$, of the form \mbox{$A
\mderives{lm} AB\alpha$}, for each derivation $D_2$ in $\Delta$,
of the form \mbox{$B \mderives{lm} b\beta$}, $b$ in $\Sigma$,
replace $D_1$ in $\Delta$ by the derivation \mbox{$A
\mderives{lm} AB\alpha \mderives{lm} Ab\beta \alpha$}. 

Here, use is made of the fact that $G$ is unambiguous and
2-tiered to write the yield of any self-embedding $A$-derivation
as $AB\alpha$. This step puts self-embedding derivations into a form
suitable to create down-shifts.

\item For each pair of derivations \mbox{$\tuple{A \mderives{lm}
\epsilon,B \mderives{lm} \alpha A\beta}$} in $\Delta^2$, with
$\alpha \neq \epsilon$, add the derivation \mbox{$B \mderives{lm}
\alpha A\beta \mderives{lm} \alpha \beta$} to $\Delta$.

After this step, each derivation into $\epsilon$ is inserted into
a derivation whose yield starts with a symbol in $\Sigma \union V$.

\item From each derivation whose yield has a leftmost terminal
symbol, build an $\epsilon$-shift and add it to $P'$.

The $\epsilon$-shift is built in the obvious way. A derivation
$D$ of the form \mbox{$A \mderives{lm} a\alpha$} is mapped to an
interpolation rule \mbox{$\tuple{A, \Gamma, r}$}, where $\Gamma$ is the
parse graph representing $D$ and $r$ is the root of~$\Gamma$.

\item From each derivation $D$ of the form
\mbox{$A\mderives{lm}Ab\alpha$}, build a down-shift
\mbox{$\tuple{A, \Gamma, t}$}, where $\Gamma$ is the parse graph
representing $D$ and $t$ is the leftmost node (labeled $A$) in
the frontier of $\Gamma$.

\end{enumerate} 

\begin{example}
Consider the 2-tiered~CFG \mbox{$\tuple{\{ A_1,A_2,B_1,B_2,
C_1,C_2,D_1,D_2,E,S \}, \{ a,b,c,d \}, P,
S}$}, with $P$ consisting of the following productions:
\begin{displaymath} \begin{array}{lll}
S & \rightarrow & A_1 B_1 \pipe C_1 E \\
A_1 & \rightarrow & A_2 \pipe \epsilon \\
A_2 & \rightarrow & a \\
B_1 & \rightarrow & B_1 B_2 \pipe B_2 \\
B_2 & \rightarrow & b \\
C_1 & \rightarrow & C_2 D_1 \\
C_2 & \rightarrow & c \\
D_1 & \rightarrow & D_1 D_2 \pipe \epsilon \\
D_2 & \rightarrow & d \\
E   & \rightarrow & \epsilon
\end{array} \end{displaymath} 

\begin{description}
\item [Step 1] The following derivations are inserted into
$\Delta$:
\begin{displaymath} \begin{array}{lll}
D_1:  & S \derives{lm} A_1 B_1 \derives{lm} A_2 B_1 \derives{lm} a B_1 \\
D_2:  & S \derives{lm} A_1 B_1 \derives{lm} B_1 \derives{lm} B_1 B_2 \\
D_3:  & S \derives{lm} A_1 B_1 \derives{lm} B_1 \derives{lm} B_2 \derives{lm} b \\
D_4:  & S \derives{lm} C_1 E \derives{lm} C_2 D_1 E \derives{lm} c D_1 E \\
D_5:  & A_1 \derives{lm} A_2 \derives{lm} a \\
D_6:  & A_1 \derives{lm} \epsilon \\
D_7:  & A_2 \derives{lm} a \\
D_8:  & B_1 \derives{lm} B_1 B_2 \\
D_9:  & B_1 \derives{lm} B_2 \derives{lm} b \\
D_{10}: & B_2 \derives{lm} b \\
D_{11}: & C_1 \derives{lm} C_2 D_1 \derives{lm} c D_1 \\
D_{12}: & C_2 \derives{lm} c \\
D_{13}: & D_1 \derives{lm} D_1 D_2 \\
D_{14}: & D_1 \derives{lm} \epsilon \\
D_{15}: & D_2 \derives{lm} d \\
D_{16}: & E \derives{lm} \epsilon
\end{array} \end{displaymath} 

Note that derivations $D_2$ and $D_3$ involve the
$\epsilon$-production for $A_1$.

\item [Step 2] The left-recursive derivations $D_8$ and $D_{13}$
are replaced by the following derivations.
\begin{displaymath} \begin{array}{lll}
{D_8}':  & B_1 \derives{lm} B_1 B_2 \derives{lm} B_1 b \\
{D_{13}}': & D_1 \derives{lm} D_1 D_2 \derives{lm} D_1 d
\end{array} \end{displaymath}

\item [Step 3] The empty derivations $D_6$, $D_{14}$ and $D_{16}$
are exhaustively plugged into the yield of other derivations in
$\Delta$, yielding eventually the following set of additional
derivations. 
\begin{displaymath} \begin{array}{lll}
D_{17}: & S \derives{lm} C_1 E \derives{lm} C_2 D_1 E \derives{lm} c D_1 E \derives{} c E \\
D_{18}: & S \derives{lm} C_1 E \derives{lm} C_2 D_1 E \derives{lm} c D_1 E \derives{} c D_1 \\
D_{19}: & S \derives{lm} C_1 E \derives{lm} C_2 D_1 E \derives{lm} c D_1 E \derives{} c E \derives{} c \\
D_{20}: & S \derives{lm} C_1 E \derives{lm} C_2 D_1 E \derives{lm} c D_1 E \derives{} c D_1 \derives{} c \\
D_{21}: & C_1 \derives{lm} C_2 D_1 \derives{lm} c D_1 \derives{lm} c \\
D_{22}: & D_1 \derives{lm} D_1 D_2 \derives{lm} D_1 d \derives{lm} d 
\end{array} \end{displaymath}

Note that $D_{19}$ and $D_{20}$ are created by plugging an
$\epsilon$-production into a derivation created during the current
step. On the other hand, the effect of \mbox{$A_1 \rightarrow
\epsilon$} is null during this step, but was taken into account
during Step~1.

\item [Step 4] All derivations in $\Delta$ except {\em (i)}~the
self-embedding derivations $D_2$, ${D_8}'$ and ${D_{13}}'$, and
{\em (ii)}~the empty derivations $D_6$, $D_{14}$, and $D_{16}$
are mapped to $\epsilon$-shifts. Note that $D_{19}$ and $D_{20}$
map to the same addendum and therefore give rise to a single rule.

\item [Step 5] The self-embedding derivations ${D_8}'$ and
${D_{13}}'$ are mapped to down-shifts.

Note that $D_2$ does not appear as such in the set of
interpolation rules $P'$, for it is redundant with the subset
\mbox{$\{ {D_8}', D_1, D_3, D_9 \}$}. On the other hand, empty
derivations are not retained either, for they are redundant with
the subset \mbox{$\{ D_2,D_3,D_{17},D_{18},D_{19},D_{20},D_{21},D_{22} \}$}.

\end{description}

\end{example}

\begin{lemma}
When the above procedure is applied to produce a GIG~$G'$ from a
CFG~$G$, all and only derivations of $G$ can be
constructed by applying rules of $G'$.
\end{lemma}

\begin{sketch} 

To prove that only derivations of $G$ are built, consider that
all addenda in the rules of $G'$ represent derivations of
$G$. Now, supposing inductively that the context to which a rule
of $G'$ applies represents a derivation of $G$, it can easily be
shown that the resulting context will then also represent a
derivation of $G$, whether the rule applied was a down-shift or
an $\epsilon$-shift.

To prove that all derivations of $G$ can be built by applying
rules of $G'$, we will first of all ignore the effect of left
recursion and $\epsilon$-productions, and then factor them back
in. We accordingly define $\mderives{lm!}$ to denote a sequence
of zero or more leftmost derivation steps that do not involve a
left-recursive cycle or $\epsilon$-production, and will
explicitly denote derivation steps that may involve an
$\epsilon$-production as $\derives{lm_{\epsilon}}$.

The proof will hinge on {\em leftmost} derivations. However, a
parse graph represents every possible derivation order;
therefore, showing that all leftmost derivations of $G$ can be
represented by parse graphs of $G'$ is sufficient to prove that
all derivations of $G$ can.

\begin{enumerate} 
\item For any leftmost derivation of $G$ of the form \mbox{$S
\mderives{lm!} xA\alpha \mderives{lm!} xa\beta\alpha$}, if we
suppose inductively that there is a GIG context that represents
\mbox{$S \mderives{lm!} xA\alpha$}, then, by
construction (Steps~1 and~4), there is an $\epsilon$-shift of
$G'$ that maps this context to one that
represents \mbox{$S \mderives{lm!}  xA\alpha \mderives{lm!}
xa\beta\alpha$}.

\item In a 2-tiered CFG, every leftmost derivation involving a
left-recursive cycle contains a subderivation of the form
\mbox{$S \mderives{lm} x_1A\alpha \pderives{lm} x_1AB\beta\alpha
\pderives{lm} x_1x_2B\beta\alpha \pderives{lm}
x_1x_2a\gamma\beta\alpha$}, with $x_1$ and $x_2$ in $\Sigma^*$,
and $a$ in $\Sigma$.

Now, supposing inductively 
\linebreak[2] there 
\linebreak[2] is a 
\linebreak[2] GIG 
\linebreak[3] context 
\linebreak[3] corresponding 
\linebreak[3] to \linebreak[3]
\mbox{$S \mderives{lm} x_1A\alpha \pderives{lm} x_1x_2\alpha$},
then there is a down-shift of $G'$ mapping it to every possible
context representing a derivation of the form 
\begin{displaymath}
S \mderives{lm} x_1A\alpha \pderives{lm!} x_1AB\beta\alpha \pderives{lm}
x_1x_2B\beta\alpha \pderives{lm!} x_1x_2a\gamma\beta\alpha
\end{displaymath}.
Now, since every elementary left-recursive cycle is captured by a
down-shift (due to exhaustive derivation down to a leftmost
cyclic occurrence or a terminal node), further application of
down-shifts can generate all possible derivations not involving
$\epsilon$-productions.

\item To prove that all cases induced by the presence of
$\epsilon$-productions are captured, it is sufficient to prove
that, if \mbox{$S \mderives{lm} xA\alpha$} is an $\epsilon$-free
derivation of $G$ and \mbox{$A \rightarrow \epsilon$} is a member
of $P$, then $G'$ can build the derivation \mbox{$S \mderives{lm}
xA\alpha \derives{lm_{\epsilon}} x\alpha$}.

If $A$ appears in the frontier of an addendum built in Steps~1
and~2, then Step~3, together with the fact that any
$\epsilon$-free derivation of $G$ can be built by $G'$,
guarantees that the effect of $A \rightarrow \epsilon$ is
captured. Otherwise, there necessarily exists a derivation
\mbox{$B \mderives{lm} A\beta$}, and therefore \mbox{$B
\mderives{lm} A\beta \derives{lm_{\epsilon}}
\beta$} will necessarily be part of a derivation produced during
Step~1. Eventually, this derivation will either be the topmost
part of an addendum or, if it is a derivation into $\epsilon$,
this derivation, $B \pderives{lm_{\epsilon}} \epsilon$, behaves
like an empty production and its effect is taken into account in
the same way as $A \rightarrow \epsilon$. Now, since we have made
the hypothesis that $G$ does not recognize the empty string,
there is no infinite regression; that is to say that \mbox{$A
\rightarrow \epsilon$} necessarily applies within a derivation \mbox{$X
\mderives{lm_{\epsilon}} \alpha B\beta \mderives{lm_{\epsilon}} \epsilon$}
such that $X$ occurs in the frontier of an addendum built during
Steps~1 and~2.

Therefore, every possible incidence of an arbitrary empty
production is captured by the procedure.

\end{enumerate} 
\end{sketch} 

If $G'$ builds the same leftmost derivations as $G$, the language
it recognizes is strongly equivalent to the language recognized
by~$G$.
\end{sketch}

\section{Deterministic GIGs and CFGs}
\label{determinism.sec}

If, for any instantaneous description $\tuple{\Gamma_1, w}$, where
$\Gamma_1$ is a GIG context whose yield is $x$ and the string to parse
is $xw$, no more than one rule is applicable, the GIG has a
deterministic behavior. This behavior is interesting for its
efficiency; for this reason, this section examines the generative
power of deterministic GIGs.

\begin{theorem}
The intersection of CFLs not containing $\epsilon$ and deterministic GILs includes the
class LL(1).
\end{theorem}

\begin{sketch} 
To start with, we will prove that an LL(1) grammar can be put
into Greibach Normal Form without losing its LL(1) property. In
order to prove this, since an LL(1) grammar contains no left
recursion, it is enough to prove the following lemma.

\begin{lemma}
Let \mbox{$G=\tuple{V,\Sigma,P,S}$} be an LL(1)~grammar. Let
\mbox{$p = A \rightarrow \alpha_1 B\alpha_2$} be a production in $P$
and \mbox{$B \rightarrow \beta_1 \mid \beta_2 \mid \ldots \mid \beta_n$} be
the set of all $B$-productions, \mbox{$B \neq A$}. Let \mbox{$G'
= \tuple{V, \Sigma, P', S}$} be obtained from $G$ by deleting
production~$p$ from $P$ and adding instead the productions
\mbox{$A \rightarrow \alpha_1 \beta_1 \alpha_2 \mid \alpha_1 \beta_2 \alpha_2 \mid \ldots
\alpha_1 \beta_n \alpha_2$}. Then $G'$ recognizes the same
language as $G$ and is LL(1).
\end{lemma}

\begin{sketch}
To adapt a recursive descent parser that parses according to $G$
to parse according to $G'$, inline all calls to the procedure $B$
inside the procedure $A$. Since these calls are nonrecursive, the
behaviour of the parser will be unaffected.
\end{sketch}

Now, a GNF grammar is LL(1) if and only if there is no more than
one $A$-production whose right-hand side begins with $a$ for
given $A$ and $a$, or else more than one symbol of lookahead
would be required to select an $A$-production. It is easy to see
that, if this condition obtains, the construction procedure
outlined in Section~\ref{inclusion.sec} produces a deterministic
GIG equivalent to the source LL(1)~grammar. Indeed, in any
instantaneous description, the anchor for any rule of the type
described in Section~\ref{inclusion.sec} is the immediate
successor of the last lexeme that is present in the frontier of
the context, so at most one node is eligible as the anchor; now,
as at most one rule corresponds to the anchor symbol, the GIG
that is constructed is deterministic.
\end{sketch}

Beyond this admittedly rather drab result, it is somewhat arduous to
characterize the class of deterministic GILs. It should be noted
that this class includes some context-sensitive languages, such as the
language analyzed in section~\ref{properInclusion.sec}, which is
recognized by a deterministic GIG.

On the other hand, one can conjecture that the intersection of
CFLs and deterministic GILs properly includes LL(1) languages from
observing the sequence of context yields obtained during a GIG
derivation.

\begin{description}
\item [Vertical $\epsilon$-shift] A vertical $\epsilon$-shift
substitutes a subgraph for a dangling node immediately following
the last lexeme of $\Gamma_1$, so it maps a context whose yield is of
the form $xA\alpha$ to a context whose yield is of the form $xb\beta
\alpha$. The same mapping can be obtained by a leftmost
derivation; and, as the decision to take it depends on one token of
lookahead, a deterministic GIG using only vertical
$\epsilon$-shifts can be emulated by an LL(1)~parser. 
\item [Horizontal $\epsilon$-shift] A horizontal $\epsilon$-shift
substitutes a subgraph for a node dominating the last lexeme, so
the new subgraph starts with an $F$-edge, as shown on
Figure~\ref{horizontalShift.fig}. The yield mapping corresponding
to a horizontal $\epsilon$-shift is of the form \mbox{$x\alpha
\Rightarrow xa\beta\alpha$}, and cannot therefore be emulated by
a leftmost derivation.

\begin{figure}[htbp]
\begin{center} ~\psfig{file=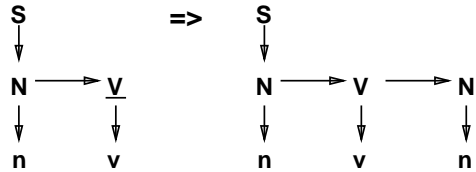,height=25mm} \end{center}
\caption{A derivation involving a horizontal $\epsilon$-shift}
\label{horizontalShift.fig}
\end{figure}

\item [Down-shift] Either the yield of the anchor is made up of
terminal symbols, as in a down-shift that encapsulates a
left-recursion, or it contains exclusively nonterminal symbols,
as in the down-shift that was used in
Section~\ref{properInclusion.sec}. In the first case, the mapping
is of the form \mbox{$xy\alpha
\Rightarrow xya\beta\alpha$}, where $y$ is the yield of the
anchor. In the second case, the mapping is of the form
\mbox{$x\alpha_1\alpha_2 \Rightarrow
xa\beta_1\alpha_1\beta_2\alpha_2$}, where $\alpha_1$ is the yield
of the anchor. Neither of these mappings can be emulated by a
leftmost derivation.
\item [Right-shift] Either the anchor is a dangling node, as in
the example that was shown on Figure~\ref{shifts.fig}, or it has
descendants, as will be the case for the rule shown on figure~\ref{rightShift.fig}. These cases
correspond respectively to \mbox{$xA\alpha \Rightarrow
xa\beta\alpha$} and \mbox{$x\alpha \Rightarrow
xa\beta\alpha$}. Only the former can be emulated by a leftmost
derivation.

\begin{figure}[htbp]
\begin{center} ~\psfig{file=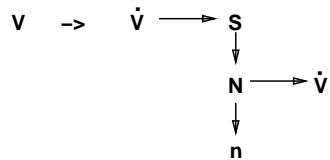,height=25mm} \end{center}
\caption{A right-shift that applies to an ancestor of the last lexeme}
\label{rightShift.fig}
\end{figure}

\end{description}

It seems possible to surmise that all yield mappings that cannot
be accounted for by left derivations, when present in a
deterministic GIG, are likely to allow it to recognize non LL(1)
languages; but what precedes can in no way be considered as a proof
of this, just as a motivation for investigating this hypothesis.

\section{Further research}

The discussion that precedes has concentrated on context-free
grammars, languages, and derivations. This can be seen as a
preliminary step to determining the amount of
context-sensitiveness supported by GIGs. In particular, the
contribution of right-shifts and horizontal $\epsilon$-shifts to
the formal power of GIGs remains to be determined.

A conclusion of this paper is that the class of deterministic
\gils\ is possibly too restricted to be satisfactory; so another
important issue for further research is the complexity of
nondeterministic GIG parsing.

\section{Related works}

It is interesting to determine how GIG-based parsing relates to
other approaches to word-by-word incremental parsing,
such as Link Grammars~\cite{Sleator95}, Dependency
Grammars~\cite{Milward92}, or Applicative Combinatory
Grammars~\cite{Milward95}. 

Its most distinctive feature with respect with these approaches
is probably that GIG parsing has as its goal the building of a
complete parse tree and each partial parse output is a
representation of a parse tree. On the contrary, parse trees do
underlie the dependency graphs produced by a link grammar or a
dependency grammar, but the parsing process does not rely on this
fact. Whether a Link Grammar or Dependency Grammar can be easily
expressed as a GIG, and with what processing consequences, is a
subject for further research.

On the other hand, incremental parsing based on Combinatory
Grammars does not generate useful phrase structures, but
incrementally builds semantic values, which constitute its real
output. Consequently, meaningful comparison with GIG-parsing
cannot be made until a semantic component is defined for GIGs.

\nocite{Hopcroft79}
\bibliography{strings,general}

\begin{thebibliography}{VSW94}

\bibitem[HU79]{Hopcroft79}
John~E. Hopcroft and Jeffrey~D. Ullman.
\newblock {\em Introduction to automata theory, languages, and computation}.
\newblock Addison-Wesley series in computer science. Addison-Wesley, 1979.

\bibitem[JLT75]{joshi-etal-75}
A.~K. Joshi, L.~S. Levy, and M.~Takahashi.
\newblock Tree adjunct grammars.
\newblock {\em Journal of Computer and System Sciences}, 10:136--63, 1975.

\bibitem[Lar98]{larcheveque98a}
John Larchev\^{e}que.
\newblock Graph interpolation grammars: a rule-based approach to the
  incremental parsing of natural languages.
\newblock Research Report RR-3390, INRIA, Rocquencourt, France, March 1998.
\newblock Available in the Computation and Language E-Print Archive under
  cmp-lg/9804001.

\bibitem[Mil92]{Milward92}
David Milward.
\newblock Dynamics, dependency grammar, and incremental interpretation.
\newblock In {\em COLING92: Proc. 14th Int. Conf. Comput. Ling., Nantes}, pages
  1095--1099, 1992.
\newblock Available from http://www.cam.sri.com/people/milward/.

\bibitem[Mil95]{Milward95}
David Milward.
\newblock Incremental interpretation of categorial grammar.
\newblock In {\em Proceedings of the 7th Conference of the European Chapter of
  the {ACL}, EACL95, Dublin}, 1995.
\newblock Available from the Computation and Language archive under
  cmp-lg/9503015.

\bibitem[ST95]{Sleator95}
Daniel D.~K. Sleator and Davy Temperley.
\newblock Parsing english with a link grammar.
\newblock Technical Report CMU-CS-TR-91-126, Carnegie Mellon University, 1995.
\newblock Available from the Computation and Language archive under
  cmp-lg/9508004.

\bibitem[VSW94]{Vijay94}
K.~Vijay-Shanker and David~J. Weir.
\newblock The equivalence of four extensions of context-free grammars.
\newblock {\em Math. Systems Theory}, 27:511--546, 1994.
\newblock Available from http://www.cogs.susx.ac.uk/lab/nlp/weir/weir.html.

\end{thebibliography}
\bibliographystyle{alpha} \end{document}